\documentclass{article}

\usepackage{microtype}
\usepackage{graphicx}
\usepackage{booktabs} 

\usepackage{amsmath, amssymb, amsthm, amsfonts, bm} 
\usepackage{caption}
\usepackage{subfig}

\usepackage{hyperref}

\usepackage[accepted]{sysml2019}

\sysmltitlerunning{Learning Index Selection with Structured Action Spaces} 

\begin{document}

\twocolumn[
\sysmltitle{Learning Index Selection with Structured Action Spaces} 




\begin{sysmlauthorlist}
\sysmlauthor{Jeremy Welborn}{cam}
\sysmlauthor{Michael Schaarschmidt}{cam}
\sysmlauthor{Eiko Yoneki}{cam}
\end{sysmlauthorlist}

\sysmlaffiliation{cam}{Computer Laboratory, University of Cambridge, Cambridge, United Kingdom}
\sysmlcorrespondingauthor{Jeremy Welborn}{jw2027@cl.cam.ac.uk}


\vskip 0.3in

\begin{abstract}
Configuration spaces for computer systems can be challenging for traditional and automatic tuning strategies. Injecting task-specific knowledge into the tuner for a task may allow for more efficient exploration of candidate configurations. We apply this idea to the task of index set selection to accelerate database workloads. Index set selection has been amenable to recent applications of vanilla deep RL, but real deployments remain out of reach. In this paper, we explore how learning index selection can be enhanced with task-specific inductive biases, specifically by encoding these inductive biases in better action structures. Index selection-specific action representations arise when the problem is reformulated in terms of permutation learning and we rely on recent work for learning RL policies on permutations. Through this approach, we build an indexing agent that is able to achieve improved indexing and validate its behavior with task-specific statistics. Early experiments reveal that our agent can find configurations that are up to 40\% smaller for the same levels of latency as compared with other approaches and indicate more intuitive indexing behavior.
\end{abstract}
]



\printAffiliationsAndNotice{}  

\section{Introduction}

Managing performance is a salient challenge in computer systems. To ensure portability in their performance, systems expose large collections of configuration parameters which must be tuned manually. Moreover, there is room for optimizations across tasks like resource allocation and scheduling. Tasks like these present as hard problems with high-dimensional configuration or action spaces. Still, they often fall to human administrators and heuristic advisory tools. 

In databases, index set selection is such a task. Index set selection seeks to speed up the execution of an expected set of queries. Choosing an optimal indexing configuration depends on complex interactions among a workload's queries, the database's distribution of data, and the database management system's query optimizer. Adding to this complexity, the space of index sets scales combinatorially.

To relieve the burden faced by human experts and heuristics across systems tasks, a range of automated strategies have been designed and deployed. Reinforcement learning (RL) is one of them and is suitable for systems tasks given its ability to bootstrap dynamic behaviors from raw signals. RL has been applied to systems tasks for over twenty years in areas like routing \cite{boyan-1994} and server resource allocation \cite{tesauro-2006} and recently, research in deep RL has featured a few applications in areas like cluster scheduling \cite{mao-2016} and TensorFlow device placement \cite{mirhoseini-2017,mirhoseini-2018}. RL can be applied offline in searching for static configurations or online (training a configuration or control policy offline and applying the policy to unseen variants of the task online). RL's ability to generalize in this way is one advantage over alternative strategies.

RL deals with sequential decision-making so can be cleanly adapted to index selection where indices may be selected in sequence across a workload. \cite{sharma-2018} and \cite{schaarschmidt-2018} have demonstrated initial deep RL controllers for simplified index selection environments. \cite{schaarschmidt-2018} construct a controller based on Deep Q-Networks \cite{mnih-2015}. Nonetheless, index set selection (like sundry combinatorial systems tasks) is characterized by complex, non-smooth, and high-dimensional state and action spaces. This exacerbates the characteristic RL challenges of trading off exploitation for exploration and assigning credit of rewards to actions taken, challenges that result in deep RL's algorithmic instability and sample inefficiency. 

We seek to add structure to the action space for the sake of finding improved indexing configurations. As articulated by works like \cite{dulac-2015,chandak-2019}, state-of-the-art RL approaches achieve sophisticated state representations (e.g. embeddings) greatly beneficial for generalization, but have not given action representation the same attention. Better representations yield better inductive biases, which we hope will enhance index set selection and in particular learning policies for index set selection. 

Our approach treats the task in part as a problem of learning permutations, and we rely on recent work for learning RL policies of permutations from \cite{emami-2018}. The Sinkhorn policy gradient algorithm can be applied straightforwardly to construct an indexing controller. Compared to early RL efforts like \cite{schaarschmidt-2018}, this approach relies on representations specifically suited for the task. The controller is able to construct improved indexing configurations for synthetic workloads based on the TPCH benchmark \cite{tpch}. In our initial experiments, our agent is able to find configurations that are up to 40\% smaller for the same levels of latency as other approaches. Its indexing behavior appears coherent and consistent with task semantics.  

To summarize, the contributions of this work are:

\begin{itemize}
	\item We demonstrate how, in an action space as complex as ours, hierarchical structure can be extracted and exploited with appropriate task-specific representations to bolster intuitive behaviors. The resulting artifact is a deep RL agent for index set selection based on Sinkhorn policy gradient algorithm.
	\item We present an evaluation of its performance based on ad hoc, intuitive statistics for indexing that show how coherent behaviors may arise out of better action structures. 
\end{itemize}

\section{Background in RL}

Reinforcement learning is a high-level approach for deriving optimal decision-making based on raw scalar reward signals. In this section, we summarize a few ideas from model-free deep reinforcement learning especially. A thorough treatment of RL can be found in \cite{sutton-barto-2018}, while surveys of deep RL are given by \cite{arulkumaran-2017,li-2017}. 

Informally, RL is appropriate whenever there is evaluative feedback available rather than instructive feedback so that behavior is updated on the basis of being constructive rather than correct \cite{sutton-barto-2018}. Formally, RL considers an agent embedded within a task environment specified as a Markov Decision Process (MDP). The agent takes actions $a \in \mathcal{A}$ to transition among states $s \in \mathcal{S}$ and in turn receives rewards over a sequence of steps or episode. At any timestep $t$, the agent takes action $a_t$ per its policy $\pi$, transitions from state $s_t$ to $s_{t+1}$, and receives reward $r_t$. In a data systems task, the state may encapsulate the current workload and configuration. The agent's goal is a policy $\pi^*$ that maximizes the expected sum of rewards or returns $\pi^* = \operatorname{argmax}_\pi \mathbb{E}_\pi[\sum_t \gamma^t r_t]$ for discount factor $\gamma$. State transitions and rewards associated with state transitions are assumed to be stochastic and Markovian, and in model-free RL these dynamics are treated as unknown. Consequently, RL agents must explore their environments thoroughly so that they may approximate this expectation with sampled experience. However, for all but small $\mathcal{S}$ and $\mathcal{A}$ spaces, the agent will have to generalize across states and actions to ensure sufficient exploration. Deep RL has received attention recently to this end. 

Model-free RL algorithms split on the basis of approximating $\pi^*$ indirectly (value-based) or directly (policy gradient). In value-based algorithms, RL learns the expected value of taking an action $a_t$ in $s_t$ as $Q^\pi(s_t,a_t)=\mathbb{E}_\pi[\sum_t \gamma^t r_t |s_t,a_t]$ and backs out an approximate $\pi^*$ by greedily selecting actions $a$ with respect to $Q^\pi(\cdot,a)$. Concretely, $Q$ is approximated with a representation $Q_\theta$, often a neural network, that can generalize from seeing a subset of $\mathcal{S} \times \mathcal{A}$. $Q_\theta$ can be set up cleverly to output $Q_\theta(s,a)$ for all $a$ in $s$, so that a single feed forward yields an action $\operatorname{argmax}_a Q_\theta(s,a)$ while decoupling state and action representations. This is termed a Deep Q-Network (DQN) \cite{mnih-2015}. $\theta$ is updated by gradient descent on $J(\theta)$, where $J(\theta)=\mathbb{E}_{s_t,a_t,r_t,s_{t+1}\sim\pi}[\frac{1}{2}(\text{target} - Q_\theta(s_t,a_t))^2]$ and $\text{target}=r_t+\gamma \max_{a_{t+1}} Q_\theta(s_{t+1},a_{t+1})$, which intuitively incorporates the ground truth given by $r_t$. 

This so-called value function approximation comes with a few complications that threaten DQN's stability. Since the training set is acquired as the agent explores, samples will be strongly correlated and targets derived from samples will suffer from non-stationarity. To address these, DQN stores samples $(s_t,a_t,r_t,s_{t+1})$ in a replay buffer and resamples batches from the buffer during SGD updates. Additionally, a set of weights is held separate to compute targets and synchronized only once in awhile with the updated weights \cite{mnih-2015}. A slew of subsequent augmentations, together termed rainbow DQN, can aid in DQN's convergence \cite{schaul-2015,van-hasselt-2016,wang-2016,hessel-2017}. 

Policy gradient algorithms instead learn $\pi^*$ as a parameterized $\pi_\theta$. $\theta$ for $\pi_\theta$ is updated directly with respect to the expected return $J(\pi_\theta)$. Somewhat surprisingly, updates in $\theta$ can be computed giving a guaranteed improvement in $J(\pi_\theta)$, despite the agent's state distribution and $\theta$'s effect on the state distribution being unknown. For stochastic $\pi$, updates are done in the direction of $\nabla_\theta \log \pi_\theta(a_t|s_t)Q^{\pi_\theta}(s_t,a_t)$ while for deterministic $\pi$, which we require, updates are done in the direction of $\nabla_\theta Q^{\pi_\theta}(s_t,\pi_\theta(s_t))$. Augmentations exist, for example, to reduce the variance of the gradient estimators given here. 

Finally, despite its advantages (e.g. online, high-dimensional decision-making in contrast to e.g. a Bayesian optimizer), systems RL faces a few challenges of its own. For example, it often takes orders of magnitude more time to interact with a system in the wild than, say, an Atari simulator so scaling to systems-sized state and action spaces becomes extremely expensive. Systems-specific RL challenges are summarized in \cite{schaarschmidt-2018}.  

\section{Learning index selection}

\subsection{Indexing semantics} 

A SQL query has to be translated from declarative SQL to an executable query plan. A query planner builds query plans based on alternative strategies for, e.g., accessing records and selects the cheapest among them according to an internal cost model. The rough aim is to reduce I/O. Indices can help here as they allow for sublinear lookup of satisfying records.

Compound indices key several attributes and can speed up a single complex query or several similar queries, saving space and time especially with clever caching. Indices over a set of attributes may serve queries having a subset of those attributes, but only if a query's attributes can be permuted into a prefix of the index attributes. This so-called prefix intersection is intuitively explained in terms of B-tree indices. B-trees are typically traversed by only one key to retrieve satisfying record addresses. However that single key can be treated as the concatenation of several keys with comparisons done in lexographic or lex order; only a prefix of the ordering will reasonably reduce the search space.

Selecting an optimal index set for a workload is non-trivial. In the best case, indexing trades query latency off with space and update time; but the impact of indexing on query latency is difficult to determine. It depends on interactions among queries, the data being queried, and the query optimizer. Index scans incur overhead, so without a real reduction in I/O using an index (versus accessing all blocks and filtering on the fly) then the optimizer may rather avoid the overhead. That reduction only occurs if a query selects for a relatively small set of records. For example, a query with an inequality constraint tends not to be as selective as a query with an equality constraint; similarly, a query with selective attributes, i.e. attributes having flatter distributions of data, tends to be a better candidate for an index scan.

\subsection{Modeling an indexing agent} 


\begin{figure}[h]
  \centering
  \includegraphics[width=1.05\columnwidth]{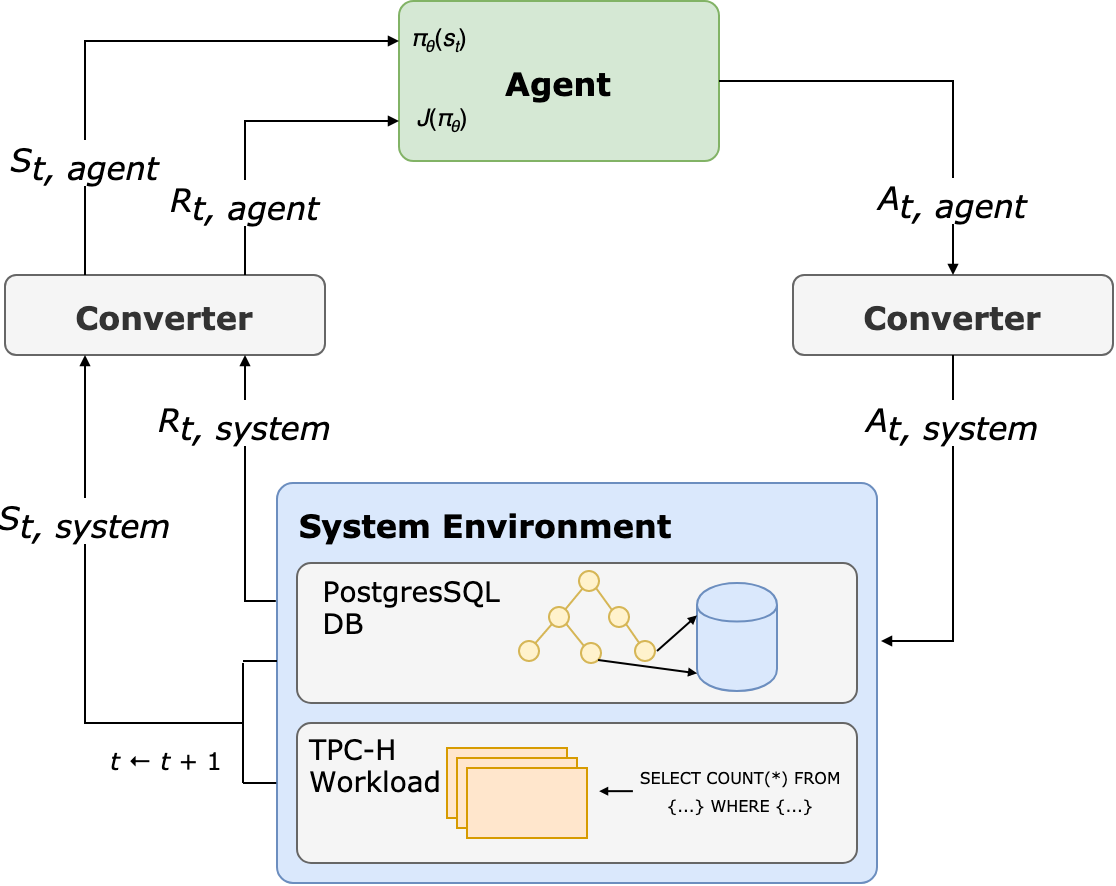}
  \caption[]{Agent-environment loop for index selection as an MDP}
  \label{fig:sys-train-loop}
\end{figure}

One interpretation of index selection as an RL task is shown in figure \ref{fig:sys-train-loop}. An agent works sequentially through a set of queries. In any timestep, the state encapsulates a query and the action encapsulates an index for that query. To take advantage of intersections, the state also stores a current index set or context, the result of actions through that timestep. For rewards, latency is the most important metric for performance, but taking into account latency alone would naively result in indices over all attributes; rather, rewards should trade off with storage space. (We ignore index update costs for convenience assuming a read-intensive, analytical workload.) Here episodes are chosen to correspond to whole workloads and during each episode, the index set is reset and rebuilt. These choices closely reflect the setup in \cite{schaarschmidt-2018}, on which our baseline DQN agent is based. Baseline representations are described in detail below with RL training summarized in algorithm \ref{alg:train}.

\textit{States:} The state represents the current query and context, and the state's representation is constrained by DQN's feedforward architecture to be a fixed-size real-valued vector. A common approach for constructing representations of arbitrary inputs is through an embedding, inspired by word embeddings \cite{mikolov-2013}. The input is tokenized, the input tokens are mapped to integers, and integers are mapped to continuous-valued vectors that are themselves trained to reflect task-relevant aspects of the tokens. As in \cite{schaarschmidt-2018}, a query \texttt{SELECT COUNT(*) FROM lineitem WHERE L\_PARTKEY = '40559' AND L\_ORDERKEY = '47914'} is tokenized. For our experiments, a simple query shape is used and shared by all inputs, so only a subset of tokens are relevant, in this case \texttt{[L\_PARTKEY', '=', 'L\_ORDERKEY', '=']}. In addition, current indices with any shared attributes are included as well. We tokenize index \texttt{['L\_SHIPINSTRUCT', 'L\_ORDERKEY']} as \texttt{['idx', 'L\_SHIPINSTRUCT\_idx', 'L\_ORDERKEY\_idx'}. The current query vector and context vector are concatenated and zero-padded as appropriate.  

\textit{Actions:} Actions specify which of a query's attributes to include in an index and, for the sake of subsequent intersection, their ordering. This suggests a few approaches for action representations. A first is a combinatorial scheme for an agent with a single output: an index is any prefix of any permutation of a query's attributes, and the agent's output is an integer that corresponds to one of these indices. In our experiments in which queries are allowed up to 3 attributes, this corresponds to $\sum_{k=0}^3 \binom{3}{k} \cdot k!=16$ actions that the agent has to disambiguate among. Scaling this up slightly to 4 attributes results in $65$ actions. 

A second is a compact scheme for an agent with several outputs. This is suggested by Schaarschmidt et al. to scale only with the number of index keys in the index. For queries with up to $n$ attributes and indices up to $m$ keys, an integer $i \in \{0,\ldots,n\}$ output from the $j \in \{1,\ldots,m\}$ output stream is treated as selecting the $i$th attribute for the $j$th key. The separate output streams can be thought of as standing for candidate index columns. Concretely, suppose $n=m=3$, the agent takes a query with query attributes \texttt{(L\_PARTKEY, L\_ORDERKEY)}, and returns $(1,0,0)$; the agent has selected the 1st query attribute \texttt{L\_PARTKEY} for the 1st index key, and no-ops for the 2nd and 3rd index keys. The advantage of this approach is its scaling, but unlike the 1st approach, not all actions in its action space are unique or even well-defined. The set of allowable actions is $\prod_{j=1}^m \{0,\ldots,n\}$. In the example, (1,0,0), (0,1,0), (1,1,0), and so on are treated as redundant, while (3,0,0) is not well-defined. 

\textit{Rewards:} In \cite{schaarschmidt-2018}, rewards are taken to be a weighted sum of space and time: $-\omega_{\text{index\_set\_size}} \cdot \text{index\_set\_size} -\omega_{\text{query\_latency}} \cdot \text{query\_latency}$. We replace $\text{index\_set\_size}$ with $\text{$\Delta$\_index\_set\_size}$, i.e. the size of an index constructed at the current step.


\begin{algorithm}[h]
	\small 
	\caption{Training algorithm for an indexing agent}
	\begin{algorithmic}
		\STATE initialize \textit{agent}, \textit{system}, \textit{converter}, \textit{workload\_builder}
		\STATE \textit{workloads} $\gets$ \textit{workload\_builder}.build()
		\FOR{\textit{workload} in \textit{workloads}}
			\STATE \textit{context} $\gets \{\}$
			\FOR{ \textit{query} in \textit{workload} }

				\STATE // get $S_t$
				\STATE \textit{agent\_state} $\gets$ \textit{converter}.to\_agent\_state(\textit{query, context})
				\vskip 4 pt
				
				\STATE // get $A_t$
				\STATE \textit{agent\_action} $\gets$ \textit{agent}.get\_action(\textit{agent\_state})
				\STATE \textit{system\_action} $\gets$ \textit{converter}.to\_system\_action(\textit{agent\_action})
				\vskip 4 pt

				\STATE // take $A_t$, get $R_t$
				\STATE \textit{index\_size} $\gets$ \textit{system}.act(\textit{system\_action})
				\STATE \textit{query\_time} $\gets$ \textit{system}.execute(\textit{query})
				\STATE \textit{agent\_reward} $\gets$ \textit{converter}.to\_agent\_reward(\textit{index\_size, query\_time})
				\vskip 4 pt

				\STATE // update agent
				\STATE \textit{agent}.observe(\textit{agent\_reward})
				\vskip 4 pt

				\STATE \textit{context}.add(\textit{system\_action})
			\ENDFOR
		\ENDFOR
	\end{algorithmic}
	\label{alg:train}
\end{algorithm}

\subsection{Limitations of a vanilla DQN agent} 

\textbf{BDQN:} We rely on the compact rather than combinatorial action representation. Thus the agent's architecture should admit several actions per index key rather than per index. A straightforward approach for this, taken in \cite{schaarschmidt-2018}, is semi-formalized in Tavakoli et al. \cite{tavakoli-2017} which introduces the idea of a branching (dueling) Q-network (BDQN). 

Whereas a DQN architecture feeds a state forward into Q-values $Q(s,a)$ over actions $a\in\mathcal{A}$, a BDQN architecture splits the top-level into streams, corresponding to Q-values $Q_d(s,a_d)$ for different action dimensions $a_d\in\mathcal{A}_d$. BDQN is trained with targets that simply average over action dimensions: $r(s,a,s^\prime) + \gamma\frac{1}{D}\sum_{d=1}^D \max_{a_d^\prime \in\mathcal{A}_d} Q_d(s^\prime, a_d^\prime)$ In \cite{tavakoli-2017}, the Reacher task is given as a good candidate for this; each of the $k$ joint angles of a $k$-jointed robot arm are actuated along different, discretized action dimensions. This avoids issues with scaling of the action space identified above. To the degree that coordination among action dimensions is desirable, the authors rely on a state representation before the split. 

\begin{figure}[h]
  \centering
  \includegraphics[width=1.1\columnwidth]{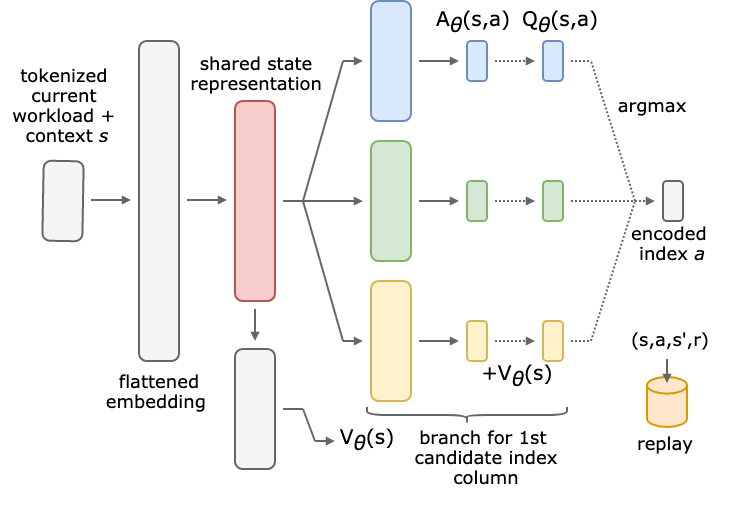}
  \caption[]{Branching DQN agent architecture. Figure adapted from \cite{tavakoli-2017}. Solid arrows correspond to fully-connected layers; dotted arrows correspond to fixed, element-wise computations. Advantage and state-action values are shown. The claim that different action dimensions are trained independently corresponds to separate weight updates and initializations over the three streams shown above.}
  \label{fig:bdqn}
\end{figure}

\textbf{Hypothetical failure cases:} The authors of \cite{tavakoli-2017} highlight how BDQN allows training over each action dimension ``semi-independently.'' It still apparently allows producing strong policies by sharing state among the branches, as shown in figure \ref{fig:bdqn}. While especially easy to implement, this approach is theoretically unmotivated. We suspect that it is suboptimal not to take directly into account the dependencies among candidate index columns or streams here. 

A few cases of $(s,a)$ tuples can help highlight the difficulty of learning indexing behavior with BDQN. In one scenario, suppose the agent receives a query with query attribute \texttt{L\_PARTKEY} and relevant context \texttt{[L\_PARTKEY, L\_ORDERKEY]}. For the agent even to explore intersection, the existence of this index has to be encoded in the shared representation and in turn in each stream so that they separately select no-ops. 

In another scenario, suppose there is no relevant context for a query with query attributes \texttt{[L\_SHIPINSTRUCT, L\_SHIPDATE]}. Suppose \texttt{L\_SHIPDATE} has so far appeared frequently throughout the workload, so that if selective, \texttt{L\_SHIPDATE} is a reasonable candidate for the first index key to increase subsequent intersections. A BDQN agent will have to figure out how cleverly to coordinate that the 1st stream selects \texttt{L\_SHIPDATE}, and the 2nd stream indexes \texttt{L\_SHIPINSTRUCT}. 

\section{Structuring the action space}

In section 3.2, index selection is seen as a sequential task with indices selected from a query's attributes. Specifically they are selected from prefixes of permutations of query attributes. In effect, we are attempting to learn a policy of permutations.

If an off-the-shelf approach like BDQN falls short of taking advantage of such task-specific structure, we are interested in representations that are able to. Indeed, representations used (for architecture, for action space) provide learning with particular inductive biases that affect how a policy generalizes from seen to unseen queries. If the representations allow easily expressing task semantics and structures, this may allow for more performant learning.

 In recent years, researchers have relied on such tailored representations in models that operate on (e.g. \cite{graves-2014}) or that optimize for (e.g. \cite{liu-2018}) discrete structures. These stem in general from gradient-admitting continuous relaxations of these structures. For problems with permutations, Mena et al. \cite{mena-2018} show how to approximate a permutation solving the assignment problem with the continuous, temperature-controlled Sinkhorn operator. And Emami et al. \cite{emami-2018} show how the Sinkhorn operator can be applied to learn permutations as RL policies, yielding the Sinkhorn Policy Gradient (SPG) algorithm. In this section, we summarize SPG and how with appropriate representations SPG can be used to construct an indexing agent. 

\subsection{Learning permutations and policies for permutations}

\textbf{Learning permutations:} A problem with learning permutations is that they are discrete and hence non-differentiable. From Birkhoff's theorem, however, permutations matrices may be readily relaxed as doubly-stochastic matrices (DSM); indeed, they are special cases of them. The Sinkhorn-Knopp algorithm allows for this relaxation by repeatedly rescaling rows and columns to sum to 1: that is, for $X\in\mathbb{R}^{N \times N}$, $S(X)$ is a DSM for $S^0(X) = \exp{(X)}, S^i(X)=\mathcal{T}_c(\mathcal{T}_r(S^{i-1}(X))), S(X)=\lim_{i\to\infty} S^i(X)$, where $\mathcal{T}_c(X)=X \oslash (\mathbf{1}_N\mathbf{1}_N^\top X)$ stands for column normalization, $\mathcal{T}_r(X)=X \oslash (X\mathbf{1}_N\mathbf{1}_N^\top)$ stands for row normalization. Of course, $S^i$ for any $i$ is differentiable \cite{sinkhorn-1967}. 

Following \cite{adams-2011}, \cite{mena-2018} adapt the so-called Sinkhorn operator $S$ in neural networks that learn permutations solving the assignment problem. In the assignment problem, workers complete tasks at a cost, and a solution seeks a cheapest, complete assignment of workers to tasks. $X$ from above may be thought of as a matrix with $X_{ij}$ as worker $i$'s cost to complete task $j$. The authors show that for temperature $\tau \to 0$, $S(X/\tau)$ approximates a solution that can be rounded cheaply by the Hungarian algorithm to allocate workers over tasks. $S(X/\tau)$ maximizes negative costs rather than minimizing costs. This can be interpreted with a great deal of generality: given an input and an evaluative signal for a permutation of the input (e.g. a reconstruction loss, a reward), a shape-conforming or square representation of the input $X$ can be learned containing likelihoods with $X_{ij}$ the likelihood of element $i$ of the input appearing in element $j$ of the input's permutation; backpropagating a signal from a proposed permutation $S(X/\tau)$ refines the representation $X$ until a proper permutation is learned.

\textbf{Learning policies of permutations:} Armed with the Sinkhorn operator, the Sinkhorn Policy Gradient algorithm learns policies of permutations with rewards rather than reconstruction error \cite{emami-2018}. The Euclidean TSP is given as a compelling candidate for this: given a candidate tour, a reward is readily available as the (negative) sum of distances between stops along the tour. 

\textit{Actor-critic setting:} The RL setup for SPG is straightforward: states $\mathcal{S}$ encapsulate a problem to be permuted, actions $\mathcal{A}$ encapsulate a permutation, and a policy (a deterministic policy) is learned $\pi : \mathcal{S} \to \mathcal{A}$. The policy gradients from section 2 adapts straightforwardly to this setting, which can be thought of as 1-step RL since the reward for, say, one TSP tour does not depend on subsequent TSP tours. Recall that for deterministic $\pi_\theta$, updates are done in the direction of $\nabla_\theta Q^{\pi_\theta}(s_t,\pi_\theta(s_t))$. Trajectories are sampled \textit{off-policy}, in this case carried out $\varepsilon$-greedily by swapping rows in action results. This allows for sample-based approximations of the gradient, so that the policy $\pi_\theta$ can be represented parametrically by a neural network, the \textit{actor}. In turn, the $Q^{\pi_\theta}(s,a)$ is approximated as $Q_{\theta^\prime}(s,a)$, the \textit{critic}.

\begin{figure*}
  \centering
  \includegraphics[scale=0.55]{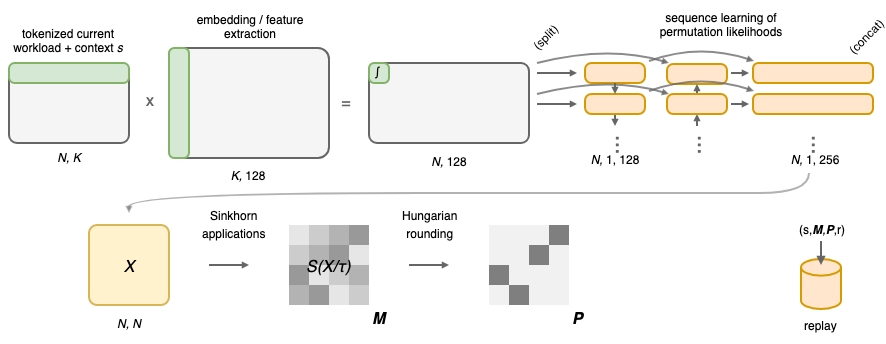}
  \caption[]{SPG actor architecture for an indexing agent. Two stages of feature extraction are emphasized: first, an embedding of the state extracts features that describe a query's attributes on their own. Second, a bidirectional-RNN extracts features that describe query attributes as they relate to one another. This is then mapped to a square matrix that admits Sinkhorn application.}
  \label{fig:spg}
\end{figure*}

\textit{Architecture:} The model for the actor used by \cite{emami-2018} is motivated by the intuitions given above: the state is of size $N \times K$; it is a sequence of size $N$ with sequence elements represented over $K$ discrete dimensions. The actor embeds or extracts features from each sequence element of the sequence, resulting in a dense higher-dimensional representation, say of size $N \times 128$. From these features, we wish to learn an $N \times N$ representation, as above. A reasonable choice for representation learning is a bi-directional RNN, allowing the actor to be roughly agnostic to ordering. We did not observe any difference among gated variants \cite{hochreiter-1997}, \cite{cho-2014}. The $N \times 128$ representation is split into a sequence of size $N$ again, fed forward through the RNN, and the $N$ outputs taken per timestep are reconcatenated. This representation can be fed through a fully-connected layer to reach the desired $N \times N$ representation. From here, the temperature-dependent Sinkhorn operator can be applied repeatedly to output an approximately desired doubly-stochastic matrix denoted $\mathbf{M}$ in \cite{emami-2018}. This can be backpropagated through, however $\mathbf{M}$ should be rounded to a sparse $\mathbf{P}$ as the action to apply. The forward computation is shown in figure \ref{fig:spg}. The critic $Q^{\pi_\theta}(s,a)$ is similar and sums embeddings of $s$ and $a$ (i.e. $\mathbf{P}$), sends this sum through a bi-directional RNN, and reduces the result to a single, scalar $Q$ value.

Since the SPG agent is off-policy, replay can alleviate the non-stationarity of targets. We propose a simple prioritization strategy for replay, resampling according to $\delta = r - Q(s,a)$ for the 1-step RL setting rather than the temporal-difference residual $\delta = r + Q(s^\prime,a^\prime) - Q(s,a)$ as in \cite{schaul-2015}.
	
\subsection{An indexing agent}

SPG's architecture is highly amenable to the indexing task. As with the baseline, we tokenize the current workload and context. Each element of the $N \times K$ state encapsulates a query attribute and information relevant to that query attribute (e.g. operators, or relevant context). This has the advantage structurally associating attributes with current, usable context. The query \texttt{SELECT COUNT(*) FROM lineitem WHERE
L\_PARTKEY = ’40559’ AND L\_ORDERKEY = ’47914’} and context \texttt{[’L\_SHIPINSTRUCT’, ’L\_ORDERKEY’]} are represented token-wise as \texttt{[['L\_PARTKEY', '=', ’idx’, ’L\_SHIPINSTRUCT\_idx’, ’L\_ORDERKEY\_idx’, 'pad', ...], ['L\_ORDERKEY', '=', 'pad', ...], ['pad', ...]]}. Unlike for a branching DQN architecture, a simple action scheme allows only well-defined indices (though redundancy is not avoided). A special \texttt{no-op} token is treated as an additional attribute of the state, so that for workloads with queries of up to 3 query attributes, $N=4$. When taking an action, a policy-asserted permutation is applied to the query attributes, but only the query attributes up to the no-op attribute are interpreted as contributing to the index. 

It is these representations from which we hypothesize SPG advantage's over DQN can be realized. For example, representation learning allows taking into account explicitly each candidate index column in relation to other candidate index columns.

\section{Evaluation}

We evaluate our agent against a DQN baseline and non-RL baselines including sophisticated random search. Our aim is to compare their performance as well as to reason about why their relative performance arises. 

\subsection{Setup and dependencies}

\textbf{PostgreSQL:} Our agents are tasked with accelerating TPC-H workloads executed against PostgreSQL, as featured in figure \ref{fig:sys-train-loop}. PostgreSQL is our database management system (DBMS) of choice; RL approaches for speeding up queries within the DBMS (i.e. query optimizers \cite{marcus-2018-a}) and outside of the DBMS (i.e. index selection \cite{sharma-2018}) have relied on PostgreSQL; the choice is consistent with other work, however not especially relevant for the research done. Here the RL agent's indexing API does not depend on PostgreSQL specifics. 

\textbf{TPCH:} The workload in turn is based on a standard benchmark, the Transaction Processing Council benchmark for ad Hoc queries or TPC-H. TPC-H works on OLAP-style workloads that comprise complex read queries that aggregate and report on stationary data from across a database. This is an appropriate setting for the indexing task because TPC-H workloads will benefit from sophisticated indexing strategies; these types of queries will be run routinely, so generalization will be advantageous. TPC-H specifies a schema encapsulating a business environment (e.g. with tables like \texttt{LINEITEM}) and a small set of 22 query templates. We assume a ``synthetic'' TPC-H workload rather than one based on the template queries, sampling queries of shape \texttt{SELECT COUNT(*) FROM ...  WHERE ...}. The \texttt{FROM} clause contains a randomly sampled TPC-H table, and the \texttt{WHERE} clause contains a randomly sampled set of constraints, with values derived from the spec. This approach is reasonable for a few reasons. For one, an agent will have a hard time extrapolating experience from 22 query templates. And while real workloads rarely have queries over only one table, earlier approaches \cite{sharma-2018,schaarschmidt-2018} work on similar simplified workloads, but it is not clear whether their agents can generalize across such simplified queries well enough to warrant scaling up the task. 

Our experiments comprise training and testing. During training, agents are run according to algorithm \ref{alg:train}. During testing, agents are exposed to workloads, set up a set of indices accordingly, and are evaluated by then running the workloads. In the results below, we train 3 DQN agents and 3 SPG agents with 100 workloads to be robust against random weight initializations and workload instantiations and test with 3 workloads. Our experiments made use of small machines (4 cores, 16 GB RAM) running PostgreSQL 9.5.17. TPC-H was set up with a scale factor of 1, and TPC-H workloads were sampled for the \texttt{LINEITEM} relation as in \cite{sharma-2018} with 25 queries and up to 3 query attribute per query. Our DQN agent implementation was based on \cite{schaarschmidt-2018} with RLGraph \cite{schaarschmidt-2018-rlgraph}. Our SPG agent was based on \cite{emami-2018,emami-2018-code}.

\subsection{Non-RL baselines}
Advisory tools for PostgreSQL are not freely available for non-enterprise setups. Instead we consider a few simple fixed baselines. One is the PostgreSQL default of primary keys, hereafter Default. Another is simple single-key indices on all attributes, hereafter Full.

In addition, random search can sometimes be a competitive alternative for deep RL \cite{salimans-2017}, \cite{mania-2018}. To this end we evaluate a search-based baseline with the tool OpenTuner \cite{ansel-2014}. OpenTuner is an autotuner with a straightforward API: a user defines a parameter space for parameters to tune and defines a run loop in which i. OpenTuner returns a configuration (a particular assignment of parameters) ii. the user executes and evaluates that configuration and iii. returns the evaluative reward to OpenTuner, which refines its search and repeats. OpenTuner searches with ensembles of search techniques (e.g. hill climbing) that are adaptive, so it is robust to different search spaces.

Our parameter space consists of a parameter per candidate index column (3 per query) per candidate index (1 per query), corresponding to the decisions carried out by our RL agents. In the run loop, we evaluate the suggested index set on a random subset of the whole training workload. The run loop runs for a fixed number of iterations, after which the top scoring configuration is saved for testing.

\subsection{Learning behaviors}

Learning is highly sensitive to hyperparametrization and we suspect pre-training like in \cite{schaarschmidt-2018} would help here, however we are concerned chiefly with relative performance in this paper. Our DQN agent uses an embedding of dimension 128, shared state of dimension 128, and streams of dimension 32. Learning relies on an Adam optimizer with learning rate of 0.001; our SPG agent uses an embedding of dimension 48, a bi-directional GRU with all dimensions of 48. Learning agent relies on an Adam optimizer with learning rates of 0.005 (actor) and 0.001 (critic). 

\begin{figure}[H]
	\captionsetup{position=top}
	\captionsetup[subfigure]{labelformat=empty} 
	\centering

	\subfloat[]{
		\includegraphics[width=\columnwidth]{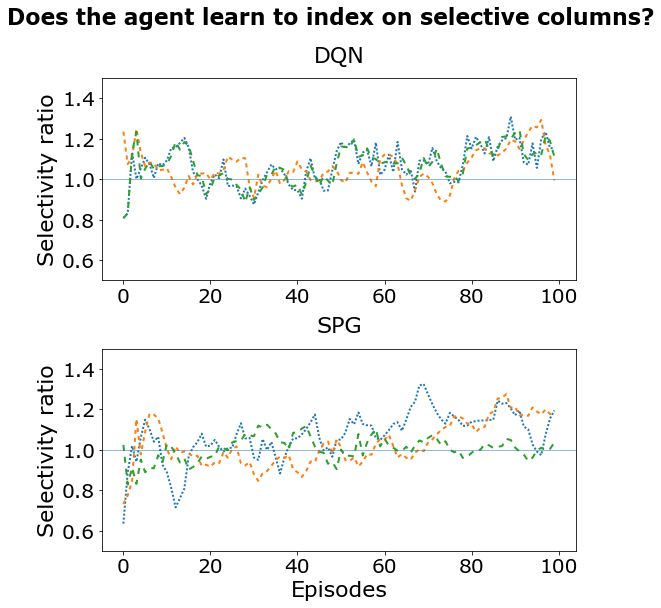}
	} \\
	\subfloat[]{
		\includegraphics[width=\columnwidth]{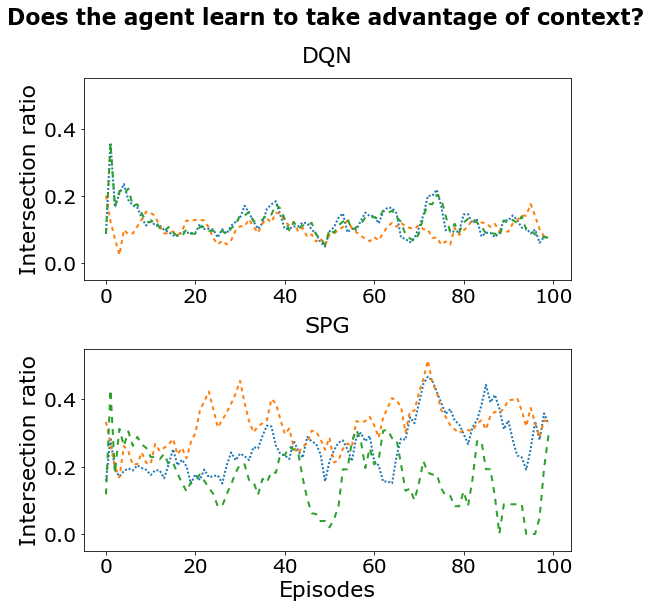}
	}
	\caption[]{Learning intuitive indexing behaviors}
	\label{fig:semantics}
\end{figure}

A surprising aspect of earlier, existing deep RL approaches like \cite{sharma-2018} and \cite{schaarschmidt-2018} is how they achieve reasonable results for index selection with such a small amount of experience. By contrast, state-of-the-art approaches for RL benchmarks like the Atari benchmark still require orders of magnitude more samples. While state and action spaces tend to be substantially bigger for those benchmarks, we still anticipate the challenge of credit assignment to be significant and sample-intensive for selecting indices. 

We want to account for this, or specifically to assess whether the behavior of DQN and SPG could be explained in terms of basic intuitions about indexing that an administrator or an advisory tool would rely on, even over a small amount of experience. The use of ad hoc, application-specific statistics is especially compelling in systems where incumbent approaches are so heavily reliant on heuristics to begin with. We ask whether the agents appear to i. learn about what comprises a ``good'' candidate index key for an index, which we can explore concretely in terms of selectivity, and whether agents appear to ii. learn complex behaviors such as relying on context, which we can explore concretely in terms of index intersections. 

\textbf{Learning about selectivity:} For i. we compute a so-called \textit{selectivity ratio} as the average selectivity of index keys in an index divided by the average selectivity of query attributes in the corresponding query. Here selectivity of an attribute is calculated as the number of unique records divided by the number of records.  

Selectivity ratios are shown throughout training of our 3 DQN and SPG agents in figure \ref{fig:semantics}. In the top subfigures, DQN and SPG have consistent selectivity curves. It seems surprised that seemingly well-performing agents achieve selectivity ratios around or slightly above 1.0. In terms of selectivity, an index with a ratio of 1 seems no better semantically than simply copying the query attributes of a query, though this does not imply necessarily that performance will be poor. A takeaway from this is that while statistics like selectivity ratio may be compelling and may even correlate with agent ability, they should not be interpreted as suggesting that a DRL agent tries to infer selectivity information from a raw reward. Interestingly, when we extended the experiment beyond 100 workloads, any additional rise in selectivity ratio did not seem to translate significantly into rewards. This could be because indexing strategies were sufficient by that time for the small task size. 

\textbf{Learning about intersections:} For ii. we relate \textit{intersection opportunities taken} to \textit{intersection opportunities}. An intersection opportunity arises if for a query, any prefix of any permutation of the query's columns is equal to any prefix of an index in the index set, which captures in particular the necessary condition for prefix intersection. We say that an intersection opportunity is taken if the agent outputs a noop or no index and if the index used by PostgreSQL instead is derived from the index considered a candidate for intersection. 

The subfigures in figure \ref{fig:semantics} below indicating intersection behavior are especially compelling. Here DQN appears not to intersect regularly and not to learn to intersect throughout training. SPG's curves suggest an exciting ability to take advantage of (and to learn to take advantage of) context. SPG's representations that encourage learning features of a query's attributes (e.g. relevant context) followed by learning features to relate those features is biased by design towards context-sensitive noops, unlike DQN.


\subsection{Performance on test workloads}

\begin{figure*}[h]
  \centering
  \includegraphics[scale=.275]{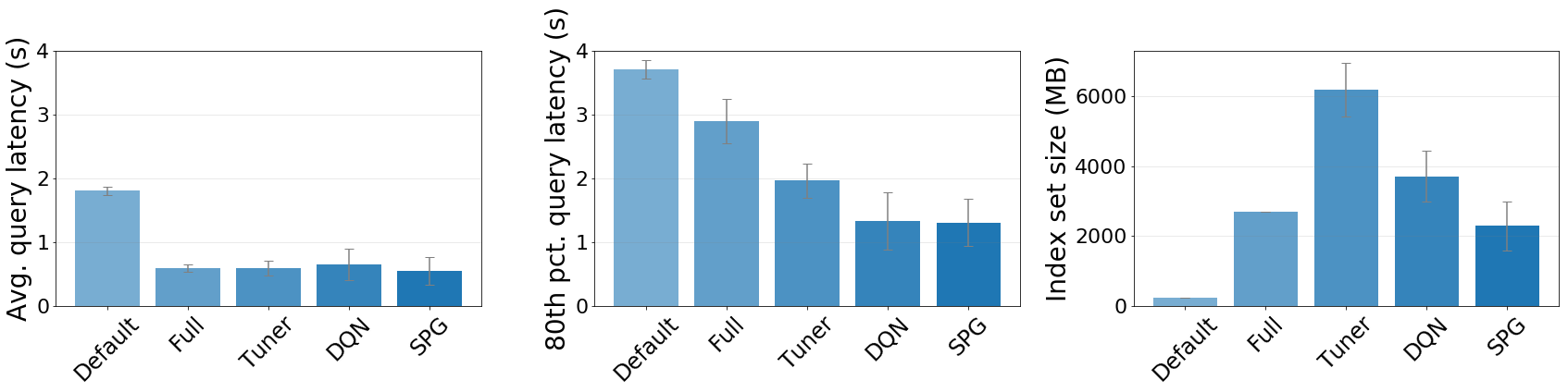}
  \caption[]{Performance evaluation}
  \label{fig:test}
\end{figure*}

To assess how these learning behaviors play out in practice, we look at performance averaged over training runs and training and test workloads. Figure \ref{fig:test} summarizes these results.  

\textbf{Characterizing workloads:} Inferences about the average test queries can be drawn by comparing mean and 80th percentile latencies. For example, consider 80th percentile latency (2.91 s) versus mean latency (0.60 s) under Full or single-key indices on all attributes. The factor of five difference suggests that single-key indices speed up queries on average, but are not able to speed up time the (at least) 5 slowest queries. This suggests that additional speedup could be achieved by compound indices. Building on this, consider 80th percentile latency (3.71 s) for Default versus 80th percentile latency (2.91s) for Full. If the difference between these is especially big, that suggests that simple indexing is sufficient for the slow queries, otherwise there is an opportunity for sophisticated indexing schemes which we hope for here.

In figure \ref{fig:test}, consider mean latencies. Full is similar to OpenTuner, DQN, and SPG. Full speeds up all queries for which a single-key index is sufficient. OpenTuner, DQN, and SPG fail on simple cases where Full succeeds, but succeed on complex cases where Full fails. Queries that Full fails on are slower than queries that OpenTuner, DQN, and SPG fail on; that mean latencies are similar across these strategies then reflects how the workload is dominated by simpler queries. Consider also 80th percentile latencies. On average, OpenTuner can identify complex indices to speed up slower queries over Full, but is beaten by DQN and SPG by 33\%. 

\textbf{Advantage over offline tuning:} There is subtlety to these results and interpreting these results. For example, it is arguably straightforward for an RL agent to speed up the slowest queries simply by constructing indices that cover all query attributes in a query and shifting any difficult decision-making to the query optimizer. This suggests why DQN and SPG agents outperform OpenTuner.  OpenTuner searches for a configuration to set up based on training queries and cannot cheaply take test queries into account, while DQN and SPG are trained to generalize from train queries so that they can cheaply set up indices for queries as queries arrive. 

\textbf{Advantages of structured action space:} \textit{Noops:} So while latency may be the salient metric for system performance, it is by trading off index set size where subtle behaviors arise. From this standpoint SPG has a strong advantage over DQN, reducing the size of the index set by almost 40\%. This is consistent with our observation that coordinating an action across action dimensions (e.g. a noop) would be challenging. On test workloads, DQN selected a noop 9.3\% of the time, while SPG took no action for a surprising 50\% of queries. On average, DQN's indices were built with 1.43 index keys versus 2.14 for SPG. The SPG agent's performance is the result of a preference for a smaller set of complex indices that can, based on the training workloads, satisfy a slew of anticipated queries.  

Additional behaviors can be characterized by inspecting individual queries and indices constructed for those queries. \textit{Query operators:} For an example of how agents respond to features of queries, consider range operators \texttt{'<'} and \texttt{'>'}. All else equal, queries with them are less likely to be executed with an index than queries without them. In the test workloads, 48\% of range queries result in noops for SPG, versus 8.7\% for DQN; similarly, 60\% of noops were taken for range queries for SPG, versus 40\% for DQN.  

\begin{figure}[H]
  \centering
  \includegraphics[width=.8\columnwidth]{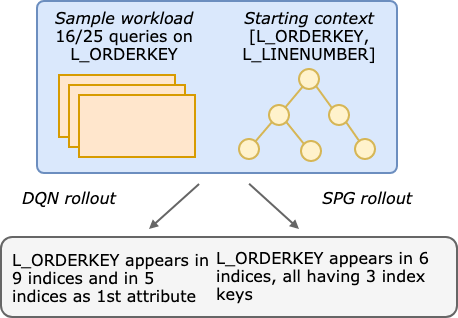}
  \caption[]{Comparing context-dependent behaviors in a sample workload.}
  \label{fig:example}
\end{figure}

\textit{Intersections:} For an example of how agents respond to features that relate queries to context, consider figure \ref{fig:example}. The \texttt{LINEITEM} attribute \texttt{L\_ORDERKEY} has high selectivity of $0.24$, so is a sensible candidate for an index column. It also appears as the 1st attribute in \texttt{LINEITEM}'s default index \texttt{[L\_ORDERKEY, L\_LINENUMBER]} however. In one of the test workloads featured in the figure, \texttt{L\_ORDERKEY} appears repeatedly in the workload. DQN tends to include it in indices that may be redundant with the default index while SPG does not. For 4 of the 5 queries that DQN indexes with \texttt{L\_ORDERKEY} as a 1st index key, the SPG agent takes no action; 3 of those queries instead intersect successfully with the default index. While it is difficult to draw certain conclusions with analyses like this, these still highlight situations in which SPG achieves intentional, intuitive indexing. 


\subsection{Summary}

Our experiments show that an SPG agent can achieve coherent indexing. At this stage they are carried out at a small scale given our restricted hardware resources and time. System evaluations, as alluded to in section 2, are expensive and in our experiments, running queries and building indices comprised on average almost $99\%$ of time per timestep. We save scaling our experiments for future work. Still, our results affirm our hypothesis that task-specific representations can be constructive in environments where samples do not come cheap, reducing best configuration sizes by around 40\%, and suggest several takeaways:

\begin{itemize}
	\item A structured action space results in fewer redundant and throwaway decisions; SPG does not have to deal with coordinating high-dimensional actions, learns better representations (e.g. for range queries), and can exploit these learned representations across a workload. 
	\item Learning can be characterized in terms of task-specific statistics; here, these reflect SPG's ability to exploit intersections over DQN.
\end{itemize}

\section{Related Work}

\textbf{Action space representations:} In an early RL effort on index set selection, \cite{sharma-2018} demonstrate a deep RL agent able to output simple or single-key indices only, simplifying the action space considerably. As discussed in section 3.1, \cite{schaarschmidt-2018} suggest an action scheme that scales with the number of allowed index keys, rather than the number of indices. 

In RL broadly, early efforts on structured action space focused on factorizing action spaces into binary subspaces; these encode actions in binary and train value functions per bit \cite{sallans-2004,pazis-2011}. In recent years, learning action embeddings (like learning state embeddings) has received attention, allowing an analogous generalization over actions \cite{vanhasselt-2009,dulac-2015,chandak-2019}; their policies specify in particular a continuous embedding space, from which a discrete action can be extracted. Our approach, by contrast, takes advantage of representations consistent with specific task semantics, e.g., as described in figure \ref{fig:spg}.

\textbf{High-dimensional configuration spaces:} A range of approaches have been proposed to deal with high-dimensional parameter spaces. One systems-specific example appears in autotuners, like OpenTuner. Autotuners evaluate the empirical performance of particular configurations to converge towards optimal ones, and some rely on Bayesian optimization to build a Gaussian process model of performance that provides the autotuner with performant candidate configurations. To alleviate the problems of parameter space, \cite{dalibard-2017} replace generic Gaussian processes with bespoke models that can be declared by system designers as probabilistic programs. Exploiting domain knowledge in this way makes exploration of candidate configurations more efficient and allows the structured auto-tuner to tackle harder, higher-dimensional tasks like structured SGD. Of course, an approach like this for combinatorial systems tasks would require additional work, e.g. constructing a custom kernel to deal with our non-smooth action space. Moreover, while a Bayesian optimizer may be sufficient in optimizing one distinct workload, it will fail to generalize and will struggle with problems beyond a few parameters. 

\section{Conclusion}
Index set selection is a task with a simple API but complex semantics. To make learning the task more tractable, in this paper we have provided an indexing agent with structured and task-specific representations of the action space. This relied on casting the problem in terms of permutation learning which lead to intuitive indexing behaviors that outperform those of an RL baseline and a non-RL, search-based autotuner. The approach is relevant for controllers and tuners across computer systems.  



\bibliography{bibliography}
\bibliographystyle{sysml2019}


\end{document}